# Exploiting Single-Cycle Symmetries in Continuous Constraint Problems

**Vicente Ruiz de Angulo**                                        RUIZ@IRI.UPC.EDU
**Carme Torras**                                              TORRAS@IRI.UPC.EDU
*Institut de Robòtica i Informàtica Industrial (CSIC-UPC)*
*Llorens i Artigas 4-6, 08028-Barcelona, Spain.*
*WWW home page:* `www.iri.upc.edu`

## Abstract

Symmetries in discrete constraint satisfaction problems have been explored and exploited in the last years, but symmetries in continuous constraint problems have not received the same attention. Here we focus on permutations of the variables consisting of one single cycle. We propose a procedure that takes advantage of these symmetries by interacting with a continuous constraint solver without interfering with it. A key concept in this procedure are the classes of symmetric boxes formed by bisecting an $n$-dimensional cube at the same point in all dimensions at the same time. We analyze these classes and quantify them as a function of the cube dimensionality. Moreover, we propose a simple algorithm to generate the representatives of all these classes for any number of variables at very high rates. A problem example from the chemical field and the cyclic $n$-roots problem are used to show the performance of the approach in practice.

## 1. Introduction

Symmetry exploitation in discrete constraint satisfaction problems (CSPs) has received a great deal of attention lately. Since CSPs are usually solved using AI search algorithms, the approaches dealing with symmetries fall into two groups: those that entail reformulating the problem or adding constraints before search (Flener, Frisch, Hnich, Kiziltan, & Miguel, 2002; Puget, 2005), and those that break symmetries along the search (Meseguer & Torras, 2001; Gent, 2002). Permutations of variables, and interchangeability of values are commonly addressed symmetries for which a repertoire of techniques have been developed, most of them relying on computational group theory.

On the contrary, symmetries have been largely disregarded in continuous constraint solving, despite the important growth in both theory and applications that this field has recently experienced (Sam-haroud & Faltings, 1996; Benhamou & Goualard, 2000; Jermann & Trombettoni, 2003; Porta, Ros, Thomas, & Torras, 2005). Continuous (or numerical) constraint solving is often tackled using Branch-and-Prune (B&P) algorithms (Hentenryck, Mcallester, & Kapur, 1997; Vu, Silaghi, Sam-Haroud, & Faltings, 2005), which iteratively locate solutions inside an initial domain box, by alternating box subdivision (branching) and box reduction (pruning) steps.

Motivated by a molecular conformation problem, in this paper we deal with the most simple type of box symmetry, namely that in which some domain variables (i.e., box dimensions) undergo a single-cycle permutation leaving the constraints invariant. To be clear, if the cycle involves $n$ variables, our algorithm handles the $n-1$ symmetries (excluding the identity) generated by this cycle by composition. Since the computational gain will be shown to be roughly proportional to $n$, the longest cycle appearing in the problem formulation should be chosen as input to our algorithm.

This single-cycle permutation that leaves the constraints unchanged is a form of constraint symmetry in the terminology introduced by Cohen, Jeavons, Jefferson, Petrie, and Smith (2006). Note that any constraint symmetry is also a solution symmetry, but not the other way around. Thus, the symmetries we deal with are a subset of all possible solution symmetries; the advantage is that they can be assessed (although perhaps are difficult to find) from the problem formulation, therefore being operative.

Our approach to exploit symmetries in continuous constraint problems requires the initial domain for the symmetric variables to be an $n$-cube, as it starts by subdividing this cube at the same point along all dimensions at once. Since box symmetry is a transitive relation, the subboxes resulting from the subdivision fall into equivalence classes. Then, a B&P algorithm (or any similar continuous constraint solver) is called on only the subboxes that are representatives of each symmetry equivalence class. Finally, for each solution found, all its symmetric ones are generated. Note that symmetry handling doesn't interfere with the inside workings of the constraint solver.

## 2. Symmetry in Continuous Constraint Problems

We are interested in solving the following general continuous Constraint Satisfaction Problem (continuous CSP): Find all points $\mathbf{x} = (x_1, \ldots, x_n)$ lying in an initial box of $\mathbb{R}^n$ satisfying the constraints $f_1(\mathbf{x}) \in C_1, \ldots, f_m(\mathbf{x}) \in C_m$, where $f_i$ is a function $f_i : \mathbb{R}^n \to \mathbb{R}$, and $C_i$ is an interval in $\mathbb{R}$.

The only particular feature that we require of a Continuous Constraint Solver (CCS) is that it has to work with an axis-aligned box in $\mathbb{R}^n$ as input. Also, we assume that the CCS returns solution boxes. Note that a CCS returning solution points is a limit case still contained in our framework.

We say that a function $s : \mathbb{R}^n \to \mathbb{R}^n$ is a point symmetry of the problem if there exists an associated permutation $\sigma \in \Sigma_m$ such that $f_i(\mathbf{x}) = f_{\sigma(i)}(s(\mathbf{x}))$ and $C_i = C_{\sigma(i)}$, $\forall i = 1, \ldots, m$. We consider symmetry as a property that relates points that are equivalent as regards to a continuous CSP. Concretely, from the above definition one can conclude that

- $\mathbf{x}$ is a solution to the problem iff $s(\mathbf{x})$ is a solution to the problem.

Let $s$ and $t$ be two symmetries of a continuous CSP with associated permutations $\sigma_s$ and $\sigma_t$. It is easy to see that the composition of symmetries $s(t(\cdot))$ is also a symmetry with associated permutation $\sigma_s(\sigma_t(\cdot))$.

An interesting type of symmetries are permutations (bijective functions of a set onto itself) of the components of $\mathbf{x}$. Let $D$ be a finite set. A cycle of length $k$ is a permutation $\psi$ such that there exist distinct elements $a_1, \ldots a_k \in D$ such that $\psi(a_i) = \psi(a_{(i+1) \bmod k})$ and $\psi(z) = z$ for any other element $z \in D$. Such a cycle is represented as $(a_1, \ldots a_k)$. Every permutation can be expressed as a composition of disjoint cycles (i.e, cycles without common elements), which is unique up to the order of the factors. Composition of cycles is represented as concatenation, as for example $(a_1, \ldots a_k)(b_1, \ldots b_l)$. In this paper we focus on a particular type of permutations, namely those constituted by a single cycle. In its simplest form[1], this is $s(x_1, x_2, \ldots x_n) = (x_{\theta(1)}, x_{\theta(2)}, \ldots x_{\theta(n)}) = (x_2, x_3 \ldots x_n, x_1)$, where $\theta(i) = (i + 1) \ mod \ n$.

**Example**: $n = 3, m = 4, \mathbf{x} = (x_1, x_2, x_3) \in [-1, 1] \times [-1, 1] \times [-1, 1]$,

$$
\begin{aligned}
f_1(\mathbf{x}) : && x_1^2 + x_2^2 + x_3^2 \in [5, 5] &\equiv\ x_1^2 + x_2^2 + x_3^2 = 5 \\
f_2(\mathbf{x}) : && 2x_1 - x_2 \in [0, \infty] &\equiv\ 2x_1 - x_2 \geqslant 0 \\
f_3(\mathbf{x}) : && 2x_2 - x_3 \in [0, \infty] &\equiv\ 2x_2 - x_3 \geqslant 0 \\
f_4(\mathbf{x}) : && 2x_3 - x_1 \in [0, \infty] &\equiv\ 2x_3 - x_1 \geqslant 0
\end{aligned}
$$

There exists a symmetry $s(x_1, x_2, x_3) = (x_2, x_3, x_1)$, for which there is no need of reordering the variables. The constraint permutation associated to $s$ is $\sigma(1) = 1$, $\sigma(2) = 3$, $\sigma(3) = 4$, and $\sigma(4) = 2$.

Generally there is not a unique symmetry for a given problem. If there exists a symmetry $s$, then for example $s^2(\mathbf{x}) = s(s(\mathbf{x}))$ is another symmetry. In general, using the convention of denoting $s^0(\mathbf{x})$ the identity mapping, $\{s^i(\mathbf{x}), i = 0 \ldots n-1\}$ is the set of different symmetries that can be obtained composing $s(\mathbf{x})$ with itself, while for $i \geqslant n$ we have that $s^i(\mathbf{x}) = s^{i \ mod \ n}(\mathbf{x})$. Thus, a single-cycle symmetry generates by composition $n - 1$ symmetries, excluding the trivial identity mapping. Some of them may have different numbers of cycles. Imagine for example that in a continuous CSP with $n = 4$ the permutation of variables (1 2 3 4) is a symmetry. Then, the permutation obtained by composing it twice, (1 3)(2 4), is also a symmetry of the problem, but has a different number of cycles, and the longest cycle has length two instead of four. Besides, the former permutation cannot be generated from the latter. The algorithm presented in this paper deals with all the compositions of a single-cycle symmetry, even if some of them are not single-cycle symmetries. The gain obtained with the proposed algorithm will be shown to be roughly proportional to the number of different compositions of the selected symmetry. Therefore, when several single-cycle symmetries exist in a continuous CSP problem, the algorithm should be used with that generating the most symmetries by composition, i.e., with that having the longest cycle. Note that the single-cycle permutations we are dealing with need not encompass all the problem variables, since the the remaining ones will be considered fixed (unitary cycles).

## 3. Box Symmetry

Since continuous constraint solvers work with boxes, we turn our attention now to the set of points symmetric to those belonging to a box $\mathcal{B} \subseteq \mathbb{R}^n$. [2]

Let $s$ be a single-cycle symmetry corresponding to the circular variable shifting $\theta$ introduced in the preceding section, and $\mathcal{B} = [\underline{x}_1, \overline{x}_1] \times \ldots \times [\underline{x}_n, \overline{x}_n]$ a box in $\mathbb{R}^n$. The *box symme-* try function $S$ is defined as $S(\mathcal{B}) = \{s(\mathbf{x}) \ \ s.t. \ \ \mathbf{x} \in \mathcal{B}\} = [\underline{x}_{\theta(1)}, \overline{x}_{\theta(1)}] \times \ldots \times [\underline{x}_{\theta(n)}, \overline{x}_{\theta(n)}] = [\underline{x}_2, \overline{x}_2] \times \ldots \times [\underline{x}_n, \overline{x}_n] \times [\underline{x}_1, \overline{x}_1]$. The box symmetry function has also an associated constraint permutation $\sigma$, which is the same associated to $s$. $S^i$ will denote $S$ composed $i$ times. We say, then, that $\mathcal{B}_1$ and $\mathcal{B}_2$ are symmetric boxes if there exists $i$ $s.t.$ $S^i(\mathcal{B}_1) = \mathcal{B}_2$.

---

1. In general, the variables must be arranged in a suitable order before one can apply the circular shifting. Thus, the general form of a single-cycle symmetry is $s(\mathbf{x}) = h^{-1}(g(h(\mathbf{x})))$, where $h(x_1, \ldots x_n) = (x_{\phi(1)}, \ldots, x_{\phi(n)})$, $\phi \in \Sigma_n$ is a general permutation that orders the variables, and $g(x_1, \ldots, x_n) = (x_{\theta(1)}, \ldots x_{\theta(n)})$ is the circular shifting above. Thus, the cycle $\psi$ defining the symmetry can be expressed as $\psi = \phi^{-1}(\theta(\phi(\cdot)))$. Since the reordering does not change substantially the presented concepts and algorithms, we have simplified the notation in the paper by assuming that the order of the component variables is the appropriate one, i.e., that $\psi = \theta$ .

2. This set $\{s(\mathbf{x}) \ \ s.t. \ \ \mathbf{x} \in \mathcal{B}\}$ is also a box if $s(\mathbf{x}) = (s_1(\mathbf{x}), \ldots, s_n(\mathbf{x})) = (g_1(x_{\phi(1)}), \ldots, g_n(x_{\phi(n)}))$, where $s_i$ is the $i$-th component of $s$, $\phi$ is an arbitrary permutation, and $g_i : \mathbb{R} \to \mathbb{R}$ is any function such that if $I$ is an interval of $\mathbb{R}$ then $\{g_i(x) \ s.t. \ x \in I\}$ is also an interval of $\mathbb{R}$.

Box symmetry is an equivalence relation defining symmetry equivalence classes. Let $R(\mathcal{B})$ be the set of *different* boxes in the symmetry class of $\mathcal{B}$, $R(\mathcal{B}) = \{S^i(\mathcal{B}), i \in \{0, \ldots, n-1\}\}$. For instance, for box $\mathcal{B}' = [0, 4] \times [2, 5] \times [2, 5] \times [0, 4] \times [2, 5] \times [2, 5]$, $R(\mathcal{B}')$ is composed of $S^0(\mathcal{B}') = \mathcal{B}'$, $S^1(\mathcal{B}') = [2, 5] \times [2, 5] \times [0, 4] \times [2, 5] \times [2, 5] \times [0, 4]$ and $S^2(\mathcal{B}') = [2, 5] \times [0, 4] \times [2, 5] \times [2, 5] \times [0, 4] \times [2, 5]$. Note that $S^3(\mathcal{B}')$ is again $\mathcal{B}'$ itself and that subsequent applications of box symmetry would repeat the same sequence of boxes. We define the *period* $P(\mathcal{B})$ of a box $\mathcal{B}$ as $P(\mathcal{B}) = |R(\mathcal{B})|$. It is easily shown that $R(\mathcal{B}) = \{S^i(\mathcal{B}), i \in \{0, \ldots, P(\mathcal{B}) - 1\}\}$. For example, for box $\mathcal{B}'$, $R(\mathcal{B}') = \{S^0(\mathcal{B}'), S^1(\mathcal{B}'), S^2(\mathcal{B}')\}$ and $P(\mathcal{B}') = 3$.

Box symmetry has implications for the continuous CSP, which are a direct consequence of the point symmetry case:

- If there is no solution inside a box $\mathcal{B}$, there is no solution inside any of its symmetric boxes either.

- A box $\mathcal{B}_f \subseteq \mathcal{B}$ is a solution iff $S^i(\mathcal{B}_f) \subseteq S^i(\mathcal{B})$ is a solution box for all $i \in \{1 \ldots P(\mathcal{B}) - 1\}$.

Sketch of proof for the first statement: Assume there is no solution inside $\mathcal{B}$ and there is some solution $x_{sol}$ inside $S^i(\mathcal{B})$. By definition of box symmetry there exists a point $x'_{sol} \in \mathcal{B}$ such that $x_{sol} = s^i(x'_{sol})$. Using the property highlighted in Section 2 we deduce that $x'_{sol}$ must be also a solution, which contradicts the hypothesis.

Sketch of proof for the second statement: A solution box is a box with at least a solution point inside. Assume $\mathcal{B}_f \subseteq \mathcal{B}$ is a solution box containing the solution point $x_{sol}$. Inside $S^i(\mathcal{B}_f)$ there is the point $s^i(x_{sol})$ that, by the property highlighted in Section 2, must be also a solution. Conversely, assume now that $S^i(\mathcal{B}_f) \subseteq \mathcal{B}$ is a solution box. Thus it contains at least a solution point, $x_{sol}$. By definition of symmetric box, this point has a symmetric point $x'_{sol} \in \mathcal{B}_f$ such that $x_{sol} = s^i(x'_{sol})$. Using the property in Section 2 again we conclude that $x'_{sol}$ must be also a solution and, thus, $\mathcal{B}_f$ is a solution box.

Both statements can be rephrased as follows :

- If the set of solution boxes contained in a box $\mathcal{B}$ is $SolSet$, the set of solution boxes contained in its symmetric box $S^i(\mathcal{B})$ is $\{S^i(\mathcal{B}_f) \ s.t. \ \mathcal{B}_f \in SolSet\}$

This means that once the solutions inside $\mathcal{B}$ have been found, the solutions inside its symmetric boxes $S^i(\mathcal{B})$, $i \in \{1 \ldots P(\mathcal{B}) - 1\}$ are available without hard calculations. In the following sections we will show how to exploit this property to save much computing time in a meta-algorithm that uses a CCS as a tool without interfering with it.

## 3.1 Box Symmetry Classes Obtained by Bisecting a $n$-cube

The algorithm we will propose to exploit box symmetry makes use of the symmetry classes formed by bisecting a $n$-dimensional cube $I^n$ (i.e., of period 1) in all dimensions at the same time and at the same point, resulting in $2^n$ boxes. We will denote $L$ and $H$ the two subintervals into which the original range $I$ is divided. For example, for $n = 2$, we have the following set of boxes $\{L \times L, L \times H, H \times L, H \times H\}$ whose periods are 1, 2, 2 and 1, respectively. And their symmetry classes are: $\{L \times L\}$, $\{L \times H, H \times L\}$, and $\{H \times H\}$. Representing the two intervals $L$ and $H$ as 0 and 1, respectively, and dropping the $\times$ symbol, the sub-boxes can be coded as binary numbers. Let $\mathbb{SR}_n$ be the set of representatives, formed by choosing the smallest box in binary order from each class. For example, $\mathbb{SR}_2 = \{00, 01, 11\}$. Note that the cube $I^n$ to be partitioned can be thought of as the the set of binary numbers of length $n$, and that $\mathbb{SR}_n$ is nothing more than a subset whose elements are different under circular shift.

The algorithm for exploiting symmetries and the way it uses $\mathbb{SR}_n$ are explained in the next section. Afterwards, in Sections 6 and 7, we study how many components $\mathbb{SR}_n$ has, how they are distributed and, more importantly, how can they be generated.

## 4. Algorithm to Exploit Box Symmetry

---

**Algorithm 1**: CSym1 algorithm.

**Input**: A $n$-cube, $[x_l, x_h] \times \cdots \times [x_l, x_h]$.
  A single-cycle box symmetry, $S$.
  A Continuous Constraint Solver, $CCS$.
**Output**: A set of boxes covering all solutions.

1  $SolutionBoxSet \leftarrow EmptySet$
2  $x^* \leftarrow \textsc{SelectBisectionPoint}(x_l, x_h)$
3  **foreach** $\mathbf{b} \in \mathbb{SR}_n$ **do**
4     $\mathcal{B} \leftarrow \textsc{GenerateSubBox}(\mathbf{b}, x_l, x_h, x^*)$
5     $SolutionBoxSet \leftarrow SolutionBoxSet \cup \textsc{ProcessRepresentative}(\mathcal{B})$
6  **return** $SolutionBoxSet$

---

The symmetry exploitation algorithm we propose uses the CCS as an external routine. The internals of the CCS must not be modified or known.

The idea is to first divide the initial box into a number of symmetry classes. Next, one needs to process only a representative of each class with the CCS. At the end, by applying box symmetries to the solution boxes obtained in this way, one would get all the solutions lying in the space covered by the whole classes, i.e., the initial box. The advantage of this procedure is that the CCS would have to process only a fraction of the initial box. Assuming that the initial box is a $n$-cube covering the same interval $[x_l, x_h]$ in all dimensions, we can directly apply the classes associated to $\mathbb{SR}_n$. A procedure to exploit single-cycle symmetries in this way is presented in Algorithm 1.

Since $\mathbb{SR}_n$ is a set of codes —not real boxes— we need a translation of the codes into boxes for the given initial box. The operator $\textsc{GenerateSubBox}(\mathbf{b}, x_l, x_h, x^*)$ returns the box $\mathcal{V} = V_1 \times \cdots \times V_n$ corresponding to code $\mathbf{b} = b_1 \ldots b_n$ when $[x_l, x_h]$ is the range of the initial box in all dimensions and $x^*$ is the point in which this interval is bisected:

$$V_i = \begin{cases} [x_l, x^*] & \text{if } b_i = 0, \\ [x^*, x_h] & \text{if } b_i = 1. \end{cases} \tag{1}$$

The point $x^*$ calculated by SELECTBISECTIONPOINT($x_l, x_h$) can be any such that $x_l < x^* < x_h$, but a reasonable one is $(x_l + x_h)/2$. The iterations over line 4 generate a set of representative boxes such that, together with their symmetries, cover the initial $n$-cube.

PROCESSREPRESENTATIVE($\mathcal{B}$) returns all the solution boxes associated to $\mathcal{B}$, that is, the solutions inside $R(\mathcal{B})$, or still in other words, the solutions inside $\mathcal{B}$ and inside its symmetric boxes. PROCESSREPRESENTATIVE($\mathcal{B}$) is based on the property stated at the end of Section 3, which allows to obtain all the solutions in the class of $\mathcal{B}$ by processing only $\mathcal{B}$ with the CCS. *SolSet* is the set of solutions found inside the representative box of the class, $\mathcal{B}$. APPLYSYMMETRY(*SolSet*, $S^i$) calculates the set of solutions of box $S^i(\mathcal{B})$ by applying $S^i$ to each of the boxes in *SolSet*. Since the number of symmetries of $\mathcal{B}$ is $P(\mathcal{B})$, the benefits of exploiting the symmetries of a class representative is proportional to its period.

---

**Algorithm 2**: The PROCESSREPRESENTATIVE function.

**Input**: A box, $\mathcal{B}$.
　　　　　A single-cycle box symmetry, $S$.
　　　　　A Continuous Constraint Solver, $CCS$.

**Output**: The set of solution boxes contained in $\mathcal{B}$ and its symmetric boxes.

1　$SolSet \leftarrow CCS(\mathcal{B})$
2　$TotalSolSet \leftarrow SolSet$
3　**for** $i=1$: $P(\mathcal{B}) - 1$ **do**
4　　$\lfloor$ $TotalSolSet \leftarrow TotalSolSet \cup$ APPLYSYMMETRY(*SolSet*, $S^i$)
5　**return** $TotalSolSet$

---

The correctness of the algorithm is easy to check. The set of boxes in which it searches explicitly or implicitly (by means of symmetry) for solutions is $\mathcal{U} = \{R(\mathcal{B})\ s.t.\ \mathcal{B}$ is a representative$\}$. In fact, $\mathcal{U}$ is the set of boxes formed by bisecting the initial box in all dimensions at the same time and at the same point. $\mathcal{U}$ covers the whole initial box and, thus, the algorithm finds *all* the solutions of the problem. Moreover, it finds each solution box only *once*, because the boxes in $\mathcal{U}$ do not have any volume in common (they share at most a "wall").

## 4.1 Discussion on the Efficiency of CSYM1

The CSYM1 algorithm launches the CCS algorithm on $|\mathbb{SR}_n|$ small boxes instead of on only the original large one. Three factors affect its efficiency as compared to that of the standard approach:

1. **Fraction of domain processed.** Only a fraction of the original domain is directly dealt with by the CCS. This fraction is a function of the periods of the $\mathbb{SR}_n$ components. One element of period $p$ represents a class formed by $p$ boxes, only one of which is processed with the CCS. Since all the boxes of the classes are of equal size, the above fraction can be calculated by dividing the number of representatives by the total number of boxes in the classes, $\frac{|\mathbb{SR}_n|}{2^n} = \frac{|\mathbb{SR}_n|}{\sum_{\mathcal{B} \in \mathbb{SR}_n} P(\mathcal{B})}$. The expected time gain is the inverse of this quantity, $\frac{\sum_{\mathcal{B} \in \mathbb{SR}_n} P(\mathcal{B})}{|\mathbb{SR}_n|}$ denoted by IFDP (Inverse of the Fraction of Domain Processed). When $n$ grows (see Section 6), the majority of the elements

of $\mathbb{SR}_n$ have period $n$, and thus IFDP tends to $n$. However, for low $n$, IFDP can be significantly smaller than $n$. This is the main factor determining the efficiency of CSym1.

2. **Smaller processed boxes.** Since the CCS initial boxes using CSym1 are $2^n$ times smaller than the original initial box, the average size of the boxes processed by the CCS is also smaller than in the standard case. Prune (box reduction or contraction) step is carried out more quickly on smaller boxes in Branch-and-Prune algorithms. In fact, best Branch-and-Prune algorithms have box contraction operators exhibiting second-order convergence, but this contraction rate requires small enough boxes to hold in practice.

3. **Number of representatives.** There is a disadvantage in fractioning excessively the initial domain. We can see this by noting that, using the original large initial box, if a contraction operator lowers the upper bound of a symmetric variable, this information could be used to lower the upper bound of the same variable in many representative boxes in $\mathbb{SR}_n$. As commented above, this contraction operator would act more strongly on the representatives themselves, but the "loss of parallelization" effect is anyway present. This factor is irrelevant for small-length cycle symmetries, say up to $n = 6$, because $|\mathbb{SR}_n|$ is very small (see Section 6 again) as compared to the number of boxes that a CCS must process in general. However, when $n$ approaches 20, the number of representatives begins to become overwhelming.

## 5. Two Illustrative Examples

The two problems below have been solved with the Branch-and-Prune CCS presented by Porta, Ros, Thomas, Corcho, Canto, and Perez (2008). It is a polytope-based method similar to that of Sherbrooke E. C. (1993) with global consistency, which exhibits quadratic convergence. The machine used to carry out all the experiments in the paper is a 2.5 Ghz G5 Apple computer.

### 5.1 Cycloheptane

Molecules can be modeled as mechanical chains by making some reasonable approximations. If two atoms are joined by a chemical bond, one can assume that there is a rigid link between them. Thus, the first approximation is that bond lengths are constant. The second one is that the angles between two consecutive bonds are also constant. In other words, the distances between the atoms in any subchain of three atoms are assumed to be constant. All configurations of the atoms of the molecule that satisfy these distance constraints, sometimes denoted *rigid-geometry hypothesis*, are valid conformations of the molecule in a kinematic sense. The constraints induced by the rigid-geometry hypothesis are particularly strong when the molecule topology forms loops, as in cycloalkanes. The problem of finding all valid conformations of a molecule can be formulated as a distance-geometry (Blumenthal, 1953) problem in which some distances between points (atoms) are fixed and known, and one must find the set of values of unknown (variable) distances that are compatible with the embedding of the points in $\mathbb{R}^3$. The unknown distances can be found by solving a set of constraints consisting of equalities or inequalities of determinants formed with subsets of the fixed and variable distances (Blumenthal, 1953).

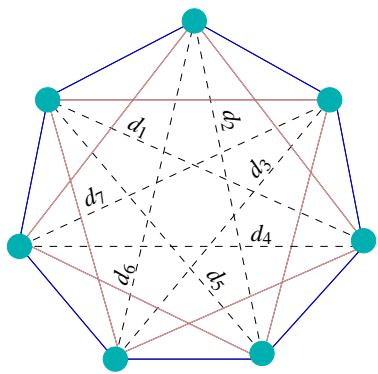

Figure 1: Cycloheptane. Disks represent carbon atoms. Constant and variable distances between atoms are represented with continuous and dashed lines, respectively.

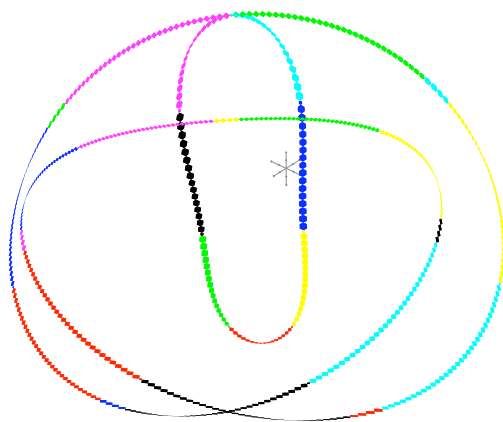

Figure 2: Three-dimensional projection of the cycloheptane solutions. The lightest (yellow) boxes are the solutions found inside the representatives using the CCS (line 1 in Algorithm 2). The other colored boxes are the solutions obtained by applying symmetries to the yellow boxes (line 4 in Algorithm 2).

Figure 1 displays the known and unknown distances of the cycloheptane, a molecule basically composed of a ring of seven carbon atoms. The distance between two consecutive atoms of the ring is constant and equal everywhere. The distance between two atoms connected to a same atom is also known and constant no matter the atoms. The problem in underconstrained, having an infinite number of solutions of dimensionality 1. The problem has several symmetries. We use one of them, $s(d_1, \ldots, d_7) = (d_{\theta(1)}, d_{\theta(2)}, \ldots, d_{\theta(7)}) = (d_2, d_3 \ldots, d_7, d_1)$. The length of the only cycle of this symmetry is $n = 7$, for which IFDP is 6.4.

The number of boxes processed using the raw CCS without symmetry handling is 1269, while using CSym1 the total number is 196, giving a ratio of $6.47 \approx$ IFDP. The problem is solved in 4.64 minutes using CSym1, which compares very favorably with the 31.6 minutes spent when using the algorithm of Porta et al. (2008) alone, a reduction by a factor of 6.81, slightly greater than IFDP. This means that, although the number of representatives begins to be relevant ($|\mathbb{SR}_7| = 20$), factor 2 in Section 4.1 is more determining than factor 3 in the same section, since the (small) time overhead introduced by handling box symmetries is also included in the reported time. Figure 2 shows a projection into $d_1$, $d_2$ and $d_3$ of the solutions obtained using CSym1. The solutions were found inside five representative boxes of period seven, containing 16, 1, 4, 64 and 1 solution boxes, respectively, at the chosen level of resolution. The total number of solutions boxes is therefore 7(16+1+4+64+1)= 602.

## 5.2 Cyclic $n$-roots Problem

The following polynomial equation system is the $n = 5$ instance of the so-called cyclic $n$-roots problem as described by Björck and Fröberg (1991).

$$
\begin{aligned}
x_1 + x_2 + x_3 + x_4 + x_5 &= 0 \\
x_1 x_2 + x_2 x_3 + x_3 x_4 + x_4 x_5 + x_5 x_1 &= 0 \\
x_1 x_2 x_3 + x_2 x_3 x_4 + x_3 x_4 x_5 + x_4 x_5 x_1 + x_5 x_1 x_2 &= 0 \\
x_1 x_2 x_3 x_4 + x_2 x_3 x_4 x_5 + x_3 x_4 x_5 x_1 + x_4 x_5 x_1 x_2 + x_5 x_1 x_2 x_3 &= 0 \\
x_1 x_2 x_3 x_4 x_5 - 1 &= 0
\end{aligned}
\tag{2}
$$

There are ten real solutions to this problem. The system has a single-cycle symmetry: $s(x_1, \ldots, x_5) = (x_2, x_3, x_4, x_5, x_1)$, as well as a multiple-cycle symmetry not considered in this paper. Thus, the cycle length is $n = 5$, $|\mathbb{SR}_5| = 8$, and the IFDP is 4. When running the CCS alone using as initial box $[-10, 10]^5$, the number of processed boxes is 399, while exploiting the aforementioned symmetry with the CSym1 algorithm this number reduces to 66. In the last case, two solutions were found in a representative box of period 5, which through symmetry led to the ten solutions. Running times are 16.86 seconds (CCS alone) and 2.08 seconds (CSym1) giving a gain of more than eight. This is the double of the IFDP, which highlights the benefits that factor 2 in Section 4.1 can bring to the efficacy of the approach. The number of representatives is very small compared to the number of boxes processed by the CCS alone, making factor 3 in Section 4.1 irrelevant in this case.

Table 1 contains the results for $n$=4 to $n$=8 of the cyclic $n$-roots problem in the $[-10, 10]^n$ domain, except for $n$=8 for which the domain was $[-5, 5]^8$. For $n$=4 and $n$=8 there is a continuum of solutions which, with the chosen resolution, produces 992 and 2435 solution boxes, respectively. Because of this, the number of processed boxes for $n$=5 is smaller than for $n$=4, but logically smaller also than for $n$=6 to $n$=8. Two observations can be made. First, the time gains are always higher than the corresponding IFDP's, implying a preponderance of factor 2 in Section 4.1 over factor 3. Second, the time gain follows rather accurately the rate between the number of processed boxes using the CCS alone and using CSYM1.

| | $n{=}4$ | $n{=}5$ | $n{=}6$ | $n{=}7$ | $n{=}8$ (reduced domain) |
|---|---|---|---|---|---|
| IFDP | 2.7 | 4.0 | 4.5 | 6.4 | 7.1 |
| number of processed boxes CCS alone | 1855 | 399 | 3343 | 38991 | 108647 |
| number of processed boxes CSYM1 | 500 | 66 | 510 | 5070 | 13304 |
| rate of processed boxes | 3.7 | 6.0 | 6.6 | 7.7 | 8.2 |
| time CCS alone | 12.0 | 16.9 | 642.0 | 20442.0 | 227355.2 |
| time CSYM1 | 3.0 | 2.1 | 95.8 | 2689.7 | 27296.5 |
| time gain CSYM1 | 4.0 | 8.1 | 6.7 | 7.6 | 8.3 |

Table 1: Results for the $n$-cyclic roots problem. Times are given in seconds.

Tests on the cyclic $n$-roots problem using a classical CCSP solver, RealPaver (Granvilliers & Benhamou, 2006), have been carried out (Jermann, 2008). The results are preliminary and difficult to expose concisely, since there is a great variability depending on issues such as the pruning method used (RealPaver offers several options) and how the problem is coded (factorized or not). In every case, however, we have observed time gains greater than expected by the IFDP.

## 6. Analysis of $\mathbb{SR}_n$: Counting the Number of Classes

Let us define some quantities of interest:

-$\mathcal{N}_n$: Number of elements of $\mathbb{SR}_n$.

-$\mathcal{FP}_n$: Number of elements of $\mathbb{SR}_n$ that correspond to full-period boxes, i.e., boxes of period $n$.

-$\mathcal{N}_{nm}$: Number of elements of $\mathbb{SR}_n$ having $m$ 1's.

-$\mathcal{FP}_{nm}$: Number of elements of $\mathbb{SR}_n$ that correspond to full-period boxes having $m$ 1's.

Polya's theorem (Polya & Read, 1987) could be used to determine some of these quantities for a given $n$ by building a possibly huge polynomial and elucidating some of its coefficients. We present a simpler way of calculating them and, at the same time, make the reader familiar with the concepts that will be used in our algorithm to generate $\mathbb{SR}_n$.

We begin by looking for the expression of $\mathcal{FP}_n$. When any number of 1's is allowed, the total number of binary numbers is $2^n$. The only periods that can exist in these binary numbers are divisors of $n$. Thus, the following equation holds:

$$\sum_{p \in div(n)} p \ \mathcal{FP}_p = 2^n. \tag{3}$$

Segregating $p = n$,

$$n \, \mathcal{FP}_n + \sum_{p \in div(n), \ p < n} p \, \mathcal{FP}_p = 2^n, \tag{4}$$

and solving for $\mathcal{FP}_n$:

$$\mathcal{FP}_n = \frac{2^n}{n} - \sum_{p \in div(n), \ p < n} \frac{p}{n} \, \mathcal{FP}_p. \tag{5}$$

This recurrence has a simple baseline condition: $\mathcal{FP}_1 = 2$.

Then, $\mathcal{N}_n$ follows easily from

$$\mathcal{N}_n = \sum_{p \in div(n)} \mathcal{FP}_p. \tag{6}$$

Segregating $p = n$, a more efficient formula is obtained:

$$\mathcal{N}_n = \frac{2^n}{n} + \sum_{p \in div(n), \ p < n} \frac{n-p}{n} \, \mathcal{FP}_p. \tag{7}$$

This formula is valid for $n > 1$. The remaining case is $\mathcal{N}_1 = 2$.

We will use similar techniques to obtain $\mathcal{FP}_{nm}$ and $\mathcal{N}_{nm}$. There are $\binom{n}{m}$ binary numbers having $m$ 1's and $n - m$ 0's. Some of these binary numbers are circular shifts of others (like 011010 and 110100). The number of shifted versions of a binary number is the period of the box being represented by the binary number. For example, 1010, of period 2, has only another shifted version, 0101. A binary number representing a box of period $p$ can be seen as a concatenation of $n/p$ numbers of length $\frac{n}{n/p} = p$ and period $p$. This means that these "concatenated" numbers are full-period, and they have $\frac{m}{n/p}$ 1's. Thus, the number of binary numbers of period $p$ when shifted numbers are counted as the same (i.e., the number of classes of period $p$) is $\mathcal{FP}_{\frac{n}{n/p} \frac{m}{n/p}}$. Only common divisors of $n$ and $m$, which we denote $div(n, m)$, can be periods. Since there are $p$ shifted versions of each binary number having period $p$, we can write

$$\sum_{p \in div(n,m)} p \, \mathcal{FP}_{\frac{n}{n/p} \frac{m}{n/p}} = \binom{n}{m}. \tag{8}$$

With a change of variable $f = n/p$ we get

$$\sum_{f \in div(n,m)} \frac{n}{f} \, \mathcal{FP}_{\frac{n}{f} \frac{m}{f}} = \binom{n}{m}. \tag{9}$$

Note that the index of the summation goes through the same values as before. We can segregate the case $f = 1$ from the summand,

$$n \, \mathcal{FP}_{nm} + \sum_{f \in div(n,m), \ f > 1} \frac{n}{f} \, \mathcal{FP}_{\frac{n}{f} \frac{m}{f}} = \binom{n}{m}, \tag{10}$$

and, finally, we obtain

$$\mathcal{FP}_{nm} = \frac{\binom{n}{m}}{n} - \sum_{f \in div(n,m), \ f>1} \frac{\mathcal{FP}_{\frac{n}{f}\frac{m}{f}}}{f}. \tag{11}$$

This is a recurrence relation from which $\mathcal{FP}_{nm}$ can be computed using the following baseline conditions:

$$\mathcal{FP}_{nn}, \mathcal{FP}_{n0} = \begin{cases} 0 & \text{if } n > 1 \\ 1 & \text{if } n = 1 \end{cases} \tag{12}$$

$\mathcal{N}_{nm}$ is obtained adding up the number of classes of each period:

$$\mathcal{N}_{nm} = \sum_{f \in div(n,m)} \mathcal{FP}_{\frac{n}{f}\frac{m}{f}}. \tag{13}$$

Segregating again $f = 1$, a more efficient formula is obtained:

$$\mathcal{N}_{nm} = \binom{n}{m} + \sum_{f \in div(n,m), \ f>1} (1 - \frac{n}{f})\mathcal{FP}_{\frac{n}{f}\frac{m}{f}}, \tag{14}$$

then carrying out the change of variable $p = n/f$:

$$\mathcal{N}_{nm} = \binom{n}{m} + \sum_{p \in div(n,m), \ p<n} (1 - p)\mathcal{FP}_{p,\frac{mp}{n}}, \tag{15}$$

Note the change in the summation range. This equation is valid whenever $m > 0$ and $m < n$. Otherwise, $\mathcal{N}_{nm} = 1$.

It is possible to extend the concept of $\mathcal{FP}_n$ (and $\mathcal{FP}_{nm}$) to reflect the number of members of $\mathbb{SR}_n$ having period $p$ (and $m$ 1's), which we denote $\mathcal{N}_n^p$ ($\mathcal{N}_{nm}^p$):

$$\mathcal{N}_n^p = \begin{cases} 0 & \text{if } p \notin div(n) \\ \mathcal{FP}_p & \text{otherwise} \end{cases} \tag{16}$$

$$\mathcal{N}_{nm}^p = \begin{cases} 0 & \text{if } p \notin div(n,m) \\ \mathcal{FP}_{p,\frac{mp}{n}} & \text{otherwise} \end{cases} \tag{17}$$

Figure 3(a) displays the number of classes ($\mathcal{N}_n$) as a function of $n$. The curve indicates an exponential-like behavior. This is confirmed in Figure 3(b) using a larger logarithmic scale, in which the curve appears almost perfectly linear. Figure 4 is an example of the distribution of classes by period for $n = 12$. Figure 5 shows the percentage of full-period classes in $\mathbb{SR}_n$ ($100\,\mathcal{N}_n^n/\mathcal{N}_n$). One can see that the percentage of classes with period different from $n$ is significant for low $n$, but approaches quickly 0 as $n$ grows. Finally, Figures 6(a) and 6(b) display the distribution of the classes in $\mathbb{SR}_n$ by number of 1's for $n = 12$ and $n = 100$, respectively. The majority of the classes concentrates in an interval in the middle of the graphic, around $n/2$. This interval becomes relatively smaller when $n$ grows.

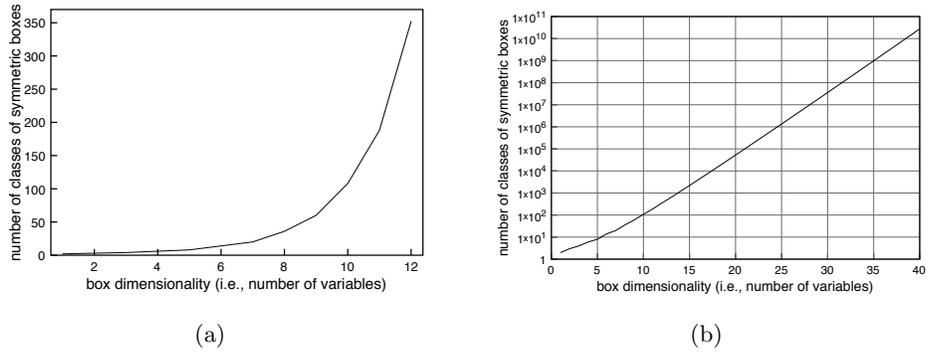

(a)            (b)

Figure 3: Number of elements of $\mathbb{SR}_n$ as a function of $n$.

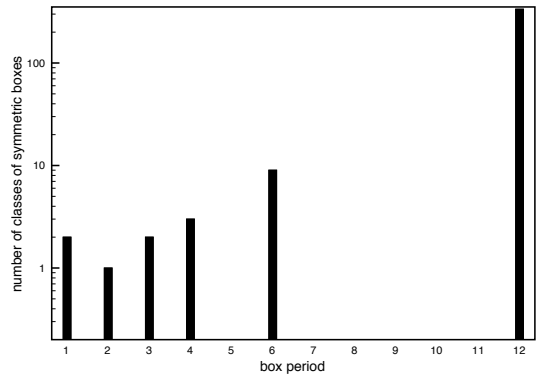

Figure 4: Number of elements of $\mathbb{SR}_{12}$ distributed by period.

## 7. Generating $\mathbb{SR}_n$, the Classes of Symmetric Boxes

The naive procedure to obtain $\mathbb{SR}_n$ would initially generate all boxes originated by bisecting a $n$-dimensional cube at the same point in all dimensions at the same time. Then, one should check each of the boxes in this set to detect whether it is a circular shift of some of the others. The complete process of generating $\mathbb{SR}_n$ in this way involves a huge number of operations even for rather small dimensions. Although the $\mathbb{SR}_n$ for a few $n$'s could be precomputed and stored in a database, we suggest here an algorithm capable of calculating $\mathbb{SR}_n$ on the fly without significant computational overhead.

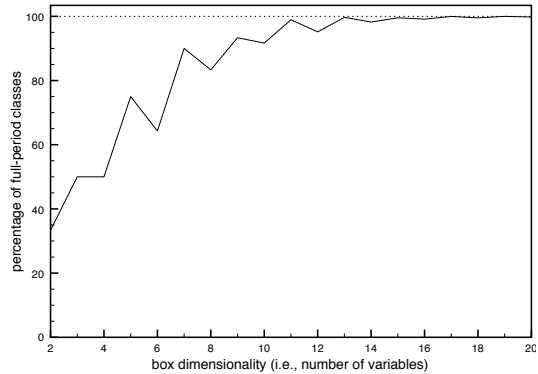

Figure 5: Percentage of full-period elements in $\mathbb{SR}_n$ as a function of $n$.

As made for counting, we distinguish different subsets of $\mathbb{SR}_n$ on the basis of the number of 1's and the period:

-$\mathbb{SR}_{nm}$: Subset of the elements of $\mathbb{SR}_n$ having $m$ 1's.

-$\mathbb{SR}_{nm}^p$: Subset of the elements of $\mathbb{SR}_n$ having $m$ 1's and period $p$.

-$\mathbb{SR}_n^p$: Subset of the elements of $\mathbb{SR}_n$ having period $p$.

From a global point of view, the generation of $\mathbb{SR}_n$ is carried out as follows. First, $\mathbb{SR}_{n0}$ is generated, which is constituted always by a unique member. Afterwards, all $\mathbb{SR}_{nm}$ for $m = 1 \ldots n$ are generated. The generation of $\mathbb{SR}_{nm}$ is divided in each of the $\mathbb{SR}_{nm}^{p}$, $p \in div(n, m)$, that compose it. The algorithm CLASSGEN described below generates all full-period representatives for any given number of variables $n > 1$ and number of ones $m > 0$, i.e., it generates $\mathbb{SR}_{nm}^{n}$. The representatives of a lower period $p \in div(n, m)$ are obtained by concatenating one same block $n/p = f$ times. Therefore, in order to obtain $\mathbb{SR}_{nm}^{p}$, we generate $\mathbb{SR}_{p\frac{m}{f}}^{p}$ with our same algorithm, and then concatenate their elements $f$ times. Thus, without loss of generality, in what follows we describe the workings of the algorithm CLASSGEN when it computes codes of full period, namely $n$.

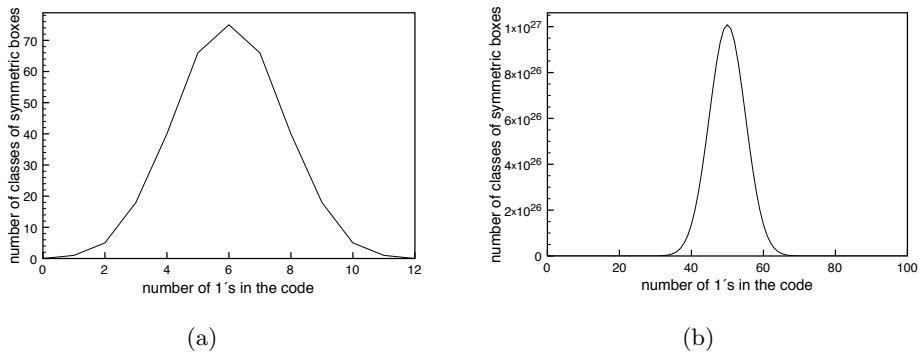

(a)                                           (b)

Figure 6: Number of elements of $\mathbb{SR}_n$ distributed by number of 1's. (a) $n$ =12. (b) $n$=100.

We use a compact coding of the binary numbers representing the boxes consisting in ordered lists or chains of numbers. The first number of the code is the number of 0's appearing before the first 1 in the binary number. The $i$-th number of the code for $i > 1$ is the number of 0's between the $(i-1)$-th and the $i$-th 1's of the binary number. For example, the number 0100010111 is codified as 13100. The length of this numerical codification is the number of 1's of the codified binary number, which has been denoted by $m$.

There are binary numbers that cannot be codified in this way, because their last digit is 0. But, except for the all zero's case, there is always an element of its class that can be codified correctly (for example 0011 is an element of the class of 0110). As our objective is to have only a representative of each class, this is rather an advantage, because half of the boxes are already eliminated from the very beginning. The all zero's box, $\mathbb{SR}_{n0}$, is common to every $n$, and will be generated separately, as already mentioned.

The codification allows to determine if a box is full-period in the same way as in the binary representation: the box has period $n$ iff after a number of circular shifts lower than the length of the numerical chain the result is never equal to the original. For instance, the example above is full-period, but 22, corresponding to 001001, is not. The only difference is that, in the new representation, at most $m$ shifts must be compared.

The code of a box can be seen as a number of base $n - m$. In a full-period box, the $m$ circular shifts of the code are different numbers, and can be arranged in strictly increasing numerical order. We will take as representative box of a class the largest element of the class when expressed as a code (which is the smallest when expressed as a binary number). For example, the class of 130 has two other elements that can be represented by our coding, 013 and 301, the latter being the chosen representative of the class.

Note that a box belonging to $\mathbb{SR}_{nm}$ has $n - m$ 0's or, equivalently, the sum of the components of the code is $n - m$.

The output of the algorithm are all codes of length $m$, whose sum of components is $n - m$, and which are both representatives of a class and full-period. Codes of length $m$ whose components sum up a desired number are rather easy to generate systematically. The representativeness and full-period conditions are more difficult to guarantee efficiently. We can handle them by exploiting the properties of our codes stated below, which make use of the definition of $i$-compability.

We say that a code is $i$-compatible or compatible for position $i$ if a sub-chain of it beginning at position $i > 1$ and ending at the last position $m$ (thus of length $m - i + 1$) is strictly smaller in numerical terms than the sub-chain of the same length beginning at the first position. For example, 423423 is compatible for positions 2 and 3, but it is not 4-compatible.

**Property 1** *A code is a class representative and it is full-period iff it is $i$-compatible for all $i$ s.t. $1 < i \leqslant m$.*

Thus, instead of comparing chains of length $m$ (i.e., the code and its shifted versions), we can determine the code validity comparing shorter sub-chains. A second property helps us to devise a still faster and simpler algorithm:

**Property 2** *If a code is $i$-compatible and the sub-chain from position $i$ to $i + l$ is equal to the sub-chain from position $1$ to $1 + l$ then the code is also compatible for positions $i + 1$ through $i + l$.*

---

**Algorithm 3**: CODEVALIDITY algorithm.

**Input**: A code of length $m$ expressed as an array, $A$.
**Output**: A boolean value indicating whether the code is valid, i.e., whether it is full-period and a class representative.

**1** $i \leftarrow 2$
**2** $ctrol \leftarrow 1$
**3** $ValidCode \leftarrow$ True
**4** **while** $ValidCode$ & $i < m$ **do**
**5**     **if** $A[i] > A[ctrol]$ **then** $ValidCode \leftarrow False$
**6**     **else if** $A[i] < A[ctrol]$ **then** $ctrol \leftarrow 1$
**7**       **else** $ctrol \leftarrow ctrol + 1$;            /* $A[i] = A[ctrol]$ */
**8**     $i \leftarrow i + 1$
**9** **if** $A[m] \geq A[ctrol]$ **then** $ValidCode \leftarrow False$
**10** **return** $ValidCode$

---

This property is interesting because it permits checking the validity of the code by travelling along it at most once, as shown in Algorithm 3. The trick is that when the decision of $i$-compatibility is being delayed because position $i$ and the following numbers are the same as those at the beginning of the string, if it finally resolves positively, the compatibility for the intermediate numbers is also guaranteed. Hence, $i$-compatibility is either resolved with a simple comparison or it requires $l$ comparisons. In the latter case, either the compatibility of $l$ positions is also resolved (if the outcome is positive) or compatibility of intermediate positions doesn't matter (because the outcome is negative and, thus, the code can be labelled non valid without further checks). A $ctrol$ variable is in charge of maintaining the last index of the "head" sub-chain that is being compared in the current compatibility check. When examining the compatibility of the current position $i$, if its value is lower than that of the $ctrol$ position, the code is for sure $i$-compatible and therefore we must only worry about $(i + 1)$-compatibility by back-warding $ctrol$ to the first position. If the value of the $ctrol$ position is equal to that of the current position $i$, the compatibility of position $i$ is still to be ascertained, and we continue advancing the current and the $ctrol$ positions until the equality disappears. In other words, the only condition that must be fulfilled for non rejecting as invalid a code at position $i$ is that

---

**Algorithm 4**: CLASSGEN algorithm.

---

**Input**: The sum of the numbers that remain to be written on the right (from position *pos* to *m*), *sum*.

The index of the next position to be written, *pos*.

The index of the current control element, whose value cannot be surpassed in the next position, *ctrol*.

The length of the code, *m*.

Array where class codes are being generated, *A*.

**Output**: A set of codes representing classes, $\mathbb{SR}$.

**1** $\mathbb{SR} \leftarrow EmptySet$

**2 if** $pos = m$ **then**

**3**     **if** $sum < A[ctrol]$ **then**        /* otherwise, $\mathbb{SR}$ will remain $EmptySet$ */

**4**        $A[m] \leftarrow sum$

**5**        $\mathbb{SR} \leftarrow \{A\};$

**6 else**

**7**     **if** $pos = 1$ **then**

**8**        $LowerLimit = \lceil sum/m \rceil$

**9**        $UpperLimit \leftarrow sum$

**10**     **else**

**11**        $LowerLimit = 0$

**12**        $UpperLimit \leftarrow$ MINIMUM$(A[ctrol], sum)$

**13**     **for** $i = UpperLimit$ **to** $LowerLimit$ **do**

**14**        $A[pos] \leftarrow i$

**15**        **if** $i = A[ctrol]$ **and** $pos \neq 1$ **then**       /* $i = A[ctrol] = UpperLimit$ */

**16**           $\mathbb{SR} \leftarrow \mathbb{SR} \bigcup$ CLASSGEN$(sum - i, pos + 1, ctrol + 1, m, A)$

**17**        **else**                    /* $i < A[ctrol]$ **or** $pos = 1$ */

**18**           $\mathbb{SR} \leftarrow \mathbb{SR} \bigcup$ CLASSGEN$(sum - i, pos + 1, 1, m, A)$

**19**

**20 return** $\mathbb{SR}$

---

$$A[i] \leqslant A[ctrol], \tag{18}$$

a condition that is transformed into $A[i] < A[ctrol]$ when $i = m$ to resolve the last of the pending compatibility checks. As an aside, note that our codes are more general than the raw binary numbers, and that representativeness and full-periodness are defined in the same way for both. Therefore, the three properties and the CODEVALIDITY algorithm apply also to the raw binary numbers.

A rather direct way to generate $\mathbb{SR}^n_{nm}$ would be to generate all the codes of length $m$ whose sum of components is $n - m$ (the number of zero's when expressed as a binary number) and then filter each of them with CODEVALIDITY. Instead, we have taken a more efficient approach, generating only the codes that satisfy the conditions that need to be checked explicitly in CODEVALIDITY. Therefore, Algorithm 3 (presented only for clarity purposes) is not used.

Our main procedure to obtain all full-period representatives having $m$ 1's, i.e., $\mathbb{SR}_{nm}^n$, is the recursive program presented in Algorithm 4. CLASSGEN$(n-m, 1, 1, m, A)$, where $A$ is an array of length $m$, must be called to obtain $\mathbb{SR}_{nm}^n$, for any given $n > 1$, $m > 0$. Each call to the procedure writes a single component of the code at the position of $A$ indicated by the parameter $pos$, beginning with $pos = 1$, which is subsequently incremented at each recursive call. The recursion finishes at the rightmost end of the code, when $pos = m$. The first parameter, $sum$, is the sum of the components of the code that remain to be written.

The range of values written at each position $pos$ is limited by $LowerLimit$ and $UpperLimit$, except for the last position $m$. In the following we show the correctness of the algorithm by verifying that these limits are chosen to satisfy the two requirements of the code:

- The sum of the numbers of any code completed by the algorithm must be $n-m$. First, recall that the initial call to the algorithm is done using a parameter $sum = n - m$. In any position $1 \leq pos < m$ the number to be written must be greater than or equal to the sum of the numbers still to be written, quantity represented by $sum$, so that in subsequent positions it will be possible to write positive integers, or at least zeros. This condition is imposed to $UpperLimit$ in line 9 for $pos = 1$ and in line 12 (juxtaposed to code validity conditions) for $1 < pos < m$. The number written at $pos$ is substracted from the $sum$ parameter in the next recursive call. Finally, for $pos = m$, the only possibility to satisfy the sum condition is to assign the value of $sum$ to the last element of the code.

- The code validity conditions, just as in CODEVALIDITY, are that the number to be written in position $pos$ must be smaller than or equal to $A[ctrol]$ for $1 < pos < m$, and strictly lower than $A[ctrol]$ for $pos = m$. These conditions are reflected in the $UpperLimit$ assignments made in lines 12 and 3, respectively. $LowerLimit$ is usually ($pos < 1$, line 4) set to the smallest possible element of the codes, 0. But at the beginning of the code ($pos = 1$, line 8) a more tight value can be chosen since, for a value lower than the upper rounded value $\lceil sum/m \rceil$, there is no way to distribute what remains of $sum$ among the other positions of the code without putting a value greater than the initial one, which would make any such code non-representative.

The maintenance of the $ctrol$ variable is similar to that within the CODEVALIDITY algorithm: if we write in $pos$ something strictly minor than $A[ctrol]$, $ctrol$ is back-warded to the first position. Otherwise, $ctrol$ is incremented by 1 for the next recursive call to write $pos + 1$.

The output of the algorithm is a list of valid codes in decreasing numerical order. For instance, the output obtained when requesting $\mathbb{SR}_{93}^9$ with CLASSGEN$(6, 1, 1, 3, A)$ is: {600, 510, 501, 420, 411, 402, 330, 321, 312}. In this example, the only case in which the recursion arrives to $pos = m$ without returning a valid code is the frustrated code 222, whose last number is not written because the code is not full-period.

Figure 7 displays quantitative results that reflect the efficiency of CLASSGEN. The dashed line accounts for the complete times required to generate all the class representatives $\mathbb{SR}_n$ for $n = 2$ to $n = 30$. It is worth noting that only $\mathbb{SR}_{30}$ requires more than a second to be entirely generated. The continuous line encodes the division of $|\mathbb{SR}_n|$ by the time required to generate $\mathbb{SR}_n$, measured in millions of class representatives generated by second. It is evident that the efficiency of CLASSGEN is very high and that it even grows slightly with $n$. This behavior shows that the dead-ends in the recursion are statistically insignificant, which proves the tightness of the bounds used to enforce the values of the code numbers.

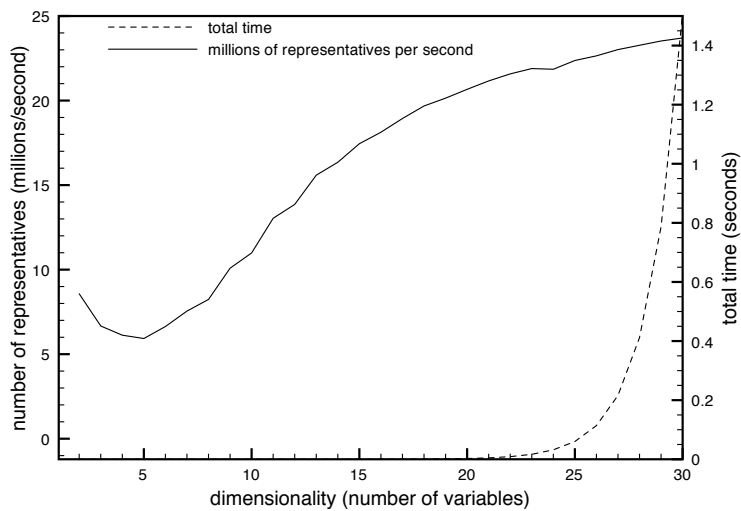

Figure 7: Total time (dashed line) to generate $\mathbb{SR}_n$, and rates of generation (continuous line) of class representatives as a function of $n$.

## 8. Conclusions

We have approached the problem of exploiting symmetries in continuous constraint satisfaction problems using continuous constraint solvers. Our approach is general and can make use of any box-oriented CCS as a black-box procedure. The particular symmetries we have tackled are single-cycle permutations of the problem variables.

The suggested strategy is to bisect the domain, the $n$-cube initial box, simultaneously in all dimensions at the same point. This forms a set of boxes that can be grouped in box symmetry classes. A representative of each class is selected to be processed by the CCS and all the symmetries of the representative are applied to the resulting solutions.

In this way, the solutions within the whole initial domain are found, while having processed only a fraction of it —the set of representatives— with the CCS. The time savings obtained by processing a representative and applying its symmetries to the solutions tend to be proportional to the number of symmetric boxes of the representative. Therefore, symmetry exploitation is complete for full-period representatives, since they have the maximum number of symmetric boxes. Another factor that improves the efficiency above what could be expected by these considerations is the smaller average size of the boxes processed by the CCS with our approach.

We have also studied the automatic generation of the classes resulting from bisecting a $n$-cube and analyzed their numerical properties. The algorithm for generating the classes is very powerful, eliminating the convenience of any pre-calculated table. The numerical analysis of the classes revealed that the average number of symmetries of the class representatives tends quickly to $n$ as the number of variables, $n$, grows. This is good news, since $n$ is the maximum number of symmetries attainable with single-cycle symmetries of $n$ variables, leading to time reductions by a factor close to $n$. Nevertheless, for small $n$ there is still a significant fraction of the representatives not having the maximum number of symmetries. Another weakness of the proposed strategy is the exponential growth in the number of classes as a function of $n$.

The problems with small and large $n$ should be tackled with a more refined subdivision of the initial domain in box symmetry classes, which is left for near future work. We are also currently approaching the extension of this work to deal with permutations of the problem variables composed of several cycles. Another complementary research line is the addition of constraints before the search with the CCS. These constraints will be specific for each symmetry class. Finally, the extension to Branch-and-Bound algorithms for nonlinear optimization could be envisaged.

## Acknowledgments

This is an extended version of work presented at CP 2007 (Ruiz de Angulo & Torras, 2007). The authors acknowledge support from the Generalitat de Catalunya under the consolidated Robotics group, the Spanish Ministry of Science and Education, under the project DPI2007-60858, and the "Comunitat de Treball dels Pirineus" under project 2006ITT-10004.